\documentclass{new_tlp}
\usepackage{mathptmx}
\usepackage{multirow}
\usepackage{lscape}
\newcommand\bcmdtab{\noindent\bgroup\tabcolsep=0pt%
  \begin{tabular}{@{}p{10pc}@{}p{20pc}@{}}}
\newcommand\ecmdtab{\end{tabular}\egroup}

\renewcommand{\theenumi}{(\roman{enumi})}

\usepackage{amsmath}
\usepackage{amssymb}

\usepackage{algorithm}
\usepackage{algorithmic}
\usepackage{xspace}
\usepackage{adjustbox}
\usepackage[normalem]{ulem}
\useunder{\uline}{\ul}{}

\usepackage{refcount}
\usepackage[justification=centering]{caption}

\usepackage{xspace}
\usepackage{multirow}
\usepackage{graphics}
\usepackage{subfigure}
\usepackage{lscape}
\usepackage{url}

\def\sr{{SR}\xspace}
\def\sri{{SRI}\xspace}
\def\srt{{SRT}\xspace}
\def\srti{{SRTI}\xspace}

\def\srtiasp{{\sc SRTI-ASP}\xspace}
\def\Personalizedsrtiasp{{Personalized-\srtiasp}\xspace}

\def\clingo{{\sc Clingo}\xspace}

\def\lar{\leftarrow}
\def\ba{\begin{array}}
\def\ea{\end{array}}
\def\beq{\begin{equation}}
\def\eeq#1{\label{#1}\end{equation}}
\def\no{\ii{not}}
\def\ii#1{\hbox{\it #1\/}}

\title[Theory and Practice of Logic Programming]
	  {Knowledge-Based Stable Roommates Problem: \\ A Real-World Application}

\author[Fidan and Erdem]{
        M\"UGE FIDAN and ESRA ERDEM\\
        Faculty of Engineering and Natural Sciences, Sabanci University, Istanbul, Turkey \\
        \{mugefidan,esra.erdem\}@sabanciuniv.edu}

\begin{document}

\maketitle
\begin{abstract}
The Stable Roommates problem with Ties and Incomplete lists (\srti) is a matching problem characterized by the preferences of agents over other agents as roommates, where the preferences may have ties or be incomplete. \srti asks for a matching that is stable and, sometimes, optimizes a domain-independent fairness criterion (e.g., Egalitarian). However, in real-world applications (e.g., assigning students as roommates at a dormitory), we usually consider a variety of domain-specific criteria depending on preferences over the habits and desires of the agents. With this motivation, we introduce a knowledge-based method to \srti considering domain-specific knowledge, and investigate its real-world application for assigning students as roommates at a university dormitory. This paper is under consideration for acceptance in Theory and Practice of Logic Programming (TPLP).
\end{abstract}



\section{Introduction}

The Stable Roommates problem (\sr)~\cite{galeShapley1962} is a matching problem characterized by the preferences of agents over other agents as roommates: each agent ranks all others in strict order of preference. A solution to \sr is then a partition of the agents into pairs that are {\em acceptable} to each other (i.e., they are in the preference lists of each other), such that the matching is {\em stable} (i.e., there exist no two agents who prefer each other to their roommates, and thus {\em block} the matching).

\sr is studied with incomplete preference lists (\sri)~\cite{gusfield89}, with preference lists including ties (\srt)~\cite{RONN1990}, and with incomplete preference lists including ties (\srti)~\cite{irving2002}. \srt and \srti are intractable under weak stability~\cite{RONN1990,Irving2009}.

Optimization variants of \srti are also studied to find more fair stable solutions. For instance, Egalitarian \srti aims to maximize the total satisfaction of preferences of all agents. Rank Maximal \srti aims to maximize the number of agents matched with their first preference, and then, subject to this condition, to maximize the number of agents matched with their second preference, and so on. Almost \srti aims to minimize the total number of blocking pairs (i.e., pairs of agents who prefer each other to their roommates), if a stable matching cannot be found. These optimization variants are NP-hard~\cite{feder1992,Cooper2020,abraham2005}.

These optimization variants of \srti are based on domain-independent measures.  However, in real-world applications (e.g., in dormitory applications), there are also domain-dependent criteria that necessitates further knowledge: consider, for instance, dormitory applications that request information about the personal habits of the students, as well as their preferences of the living environment.

In our earlier work~\cite{erdem2020}, we have developed a formal framework, called~\srtiasp, that is flexible enough to provide solutions to all variations of \sr mentioned above, including the intractable decision/optimization versions: \srt, \srti, Egalitarian \srti, Rank Maximal \srti, Almost \srti.  \srtiasp utilizes the expressive languages and efficient solvers of Answer Set Programming (ASP)~\cite{Niemelae99,MarekT99,Lifschitz02,BrewkaEL16} based on answer set semantics~\cite{GelfondL88,GelfondL91}.

In this study, we extend \srtiasp to accommodate additional domain-specific criteria in two ways: Personalized-\srti and Most-\srti. In addition, we extend \srtiasp to accommodate diversity preferences and constraints.
\begin{itemize}
\item
For Personalized-\srti, we introduce a new type of preference ordering considering (i) the importance of each criterion for each agent (e.g., one student may give more importance to sleeping habits whereas another student may give more importance to smoking habits), and (ii) the agents' preferred choices for each domain-specific criterion (e.g., whether a student prefers a roommate who does not smoke). We define an extended preference list for each agent, that combines two types of preference lists: a preference list of the agent over other agents (as in \srti) and this new type of criteria-based personalized preference list of the agent. Personalized-\srti considers these extended preference lists to compute personalized stable matchings.
\item
For Most-\srti, we introduce a new incremental definition of a stable matching considering (i) the ordering of the most preferred criteria (e.g., identified by large surveys) and (ii) the agents' preferred choices for each domain-specific criterion, with the motivation that the agents with close choices are matched. Most-\srti aims to compute such most preferred criteria based stable matchings, by utilizing the weak constraints of ASP.
\item
In addition to the students' preferences over a set of domain-specific criteria, the schools may prefer matchings (or put constraints over matchings) to increase diversity. For example, they may want to match students from different departments, classes, or countries. With this motivation, we extend \srtiasp by representing such diversity preferences/constraints using weak/hard constraints of ASP.
\end{itemize}

We illustrate a real-world application of \srtiasp by interacting with at least 200 students at Sabanci University over four surveys: (i) to decide which domain-specific criteria to consider, (ii) to collect the students preferences for domain-specific criteria, and (iii) to evaluate the usefulness of our method.

We also present results of our experiments with objective and subjective measures, to understand the scalability of the proposed two methods, Personalized-\srti and Most-\srti.


\section{\srti: Stable Roommates problem with Ties and Incomplete Lists}
{\label{sec:srti}}

We define \srti as in~\cite{erdem2020}. Let $A$ be a finite set of agents. For every agent $x\in A$, let
$A_x$ of $A \backslash \{x\}$ be a set of agents that are {\em acceptable} to $x$ as roommates. For every $y$ in $A_x$, we assume that $x$ prefers $y$ as a roommate compared to being single.

Let  $\prec_{x}$ be a partial ordering of $x$'s preferences over $A_x$ where incomparability is transitive. We refer to $\prec_{x}$ as agent $x$'s preference list. For two agents $y$ and $z$ in $A_x$, we denote by $y \prec_{x} z$ that $x$ prefers $y$ to $z$. In this context, ties correspond to indifference in the preference lists: an agent $x$ is {\em indifferent} between the agents $y$ and $z$, denoted by $y \sim_{x} z$, if $y \not \prec_{x} z$ and $z \not \prec_{x} y$. We denote by $\prec$ the collection of all preference lists.

A \emph{matching} for a given \sri instance is a function
${M: A \mapsto A}$ such that, for all $\{x,y\} \subseteq A$ such that $x\in A_y$ and $y\in A_x$, $M(x)=y$ if and only if $M(y)=x$. If agent $x$ is mapped to itself, we then say he/she is \emph{single}.

A matching $M$ is {\em blocked} by a pair $\{x, y\} \subseteq A$ ($x\neq y$) if 
\begin{itemize}
	\item[B1] both agents $x$ and $y$ are acceptable to each other,
    \item[B2] $x$ is single with respect to $M$, or $y \prec_{x} M(x)$, and
    \item[B3] $y$ is single with respect to $M$, or $x \prec_{y} M(y)$.
\end{itemize}
A matching for \srti is called {\em stable} if it is not blocked by any pair of agents.

We can declaratively solve \srti using ASP as described in~\cite{erdem2020}. For that, the input $I=(A,\prec)$ of an \srti instance is formalized by a set $F_I$ of facts using atoms of the forms $\ii{agent}(x)$ (``$x$ is an agent in $A$'') and $\ii{prefer2}(x,y,z)$ (``agent $x$ prefers agent $y$ to agent $z$, i.e., $y \prec_{x} z$''). For every agent $x$,  for every $y\in A_x$, we also add facts of the form $\ii{prefer2}(x,y,x)$ to express that $x$ prefers $y$ as a roommate instead of being single.

Based on the preferences of agents, for each agent, the concept of acceptability is defined:
$$ 
\ba l
\ii{accept}(x,y) \lar \ii{prefer}(x,y,\_). \\
\ii{accept}(x,y) \lar \ii{prefer}(x,\_,y).
\ea
$$
as well as the concept of mutual acceptability:
$$ 
\ii{accept2}(x,y) \lar \ii{accept}(x,y), \ii{accept}(y,x).
$$

The output $M: A \mapsto A$ of an \srti instance is characterized by atoms of the form $\ii{room}(x,y)$ (``agents $x$ and $y$ are roommates''). The ASP formulation $P$ of \srti first generates pairs of roommates. For every agent $x$, exactly one mutual acceptable agent $y$ is nondeterministically chosen as $M(x)$ by the choice rules:
$$ 
\ba l
1\{\ii{room}(x,y){:} \ii{agent}(y), \ii{accept2}(x,y)\}1 \lar \ii{agent}(x). \\
\lar \ii{room}(x,y), \no\ \ii{room}(y,x).
\ea
$$ 
Then, the stability of the generated matching is ensured by the hard constraints:
$$
\lar \ii{block}(x,y) \qquad (x\neq y) .
$$ 
\noindent where atoms of the form $\ii{block}(x,y)$ describe the blocking pairs (i.e., conditions B1--B3).


\section{Personalized-\srti: \srti with Personalized Criteria}

Some universities and colleges send questionnaires to students before making roommate matches, and they match students as roommates taking into account the additional information included in these questionnaires. For instance, Table~\ref{tab:form} shows a questionnaire used for applying to dormitories of the University of North Texas.\footnote{\label{northTexas}\url{https://tams.unt.edu/studentlife/roommate-preferences-questionnaire}} It contains questions about the sleep preferences, music preferences, and sharing preferences of applicants.
In some other surveys, we can see a question about the smoking habits,\footnote{\label{cleveland}\url{https://my.clevelandclinic.org/-/scassets/files/org/professionals/student-housing/roommate-questionnaire-worksheet}} preferences regarding the room temperature,\footnote{\label{wells}\url{https://www.wells.edu/files/public/forms/Housing_Roommate_Questionnaire-fillable.pdf}}  and willingness to share a room with an international student.\footnote{\label{clark}\url{https://college.lclark.edu/live/files/27111-2019-20-returning-student-questionnaire}}
In addition to questions about such different criteria, these questionnaires usually request for the applicants to indicate the most important criteria, as seen in Table~\ref{tab:form}.
\begin{table}[t]
	\caption{A dormitory application form.}
	\label{tab:form}
\vspace{-.3\baselineskip}
	\begin{minipage}{\textwidth}
	\centering
	\resizebox{0.8\textwidth}{!}{\begin{tabular}{ll}
		\hline\hline
		First Name : & \\
		Last Name : & \\
		Email : & \\
		\hline
		Gender & $\circ$ Female  $\circ$ Male\\
		\hline
		\textbf{Sleep Preferences} & \\
		$\bullet$ I understand that curfew is at 11pm Sunday-Thursday. &  $\circ$ Before 11pm \\
		 I prefer a roommate who goes to bed: * & $\circ$ Before Midnight \\
		 &  $\circ$ After Midnight \\
		 & \\
		 & $\circ$ Doesn't matter\\
		 $\bullet$ \text{ In order for me to go to sleep: *} & $\circ$ The room has to be dark \\
		& $\circ$ At least one light must be on \\
		& $\circ$ Blinds open to let the morning sun in \\
		& $\circ$ Doesn't matter \\& \\
		$\bullet$ \text{ I absolutely can NOT go to sleep: *} & $\circ$ Unless it is absolutely quiet \\
		& $\circ$ With the TV or stereo on  \\
		& $\circ$ If my roommate has visitors in the room  \\
		& $\circ$ I can sleep through anything\\
		\hline
		\textbf{Music Preferences} & \\
		$\bullet$ Favourite Type of Music & $\circ$ Rock $\circ$ Hip-Hop $\circ$ Classical $\circ$ Indie \\
		& $\circ$ Pop $\circ$ Rap $\circ$ Electronic $\circ$ Religious \\
		& \\
		$\bullet$ Least Favourite Type of Music & $\circ$ Rock $\circ$ Hip-Hop $\circ$ Classical $\circ$ Indie \\
		& $\circ$ Pop $\circ$ Rap $\circ$ Electronic $\circ$ Religious \\
		\hline
		\textbf{Sharing Preferences} &  \\	
		$\bullet$ I tend to keep my room/personal space and belongings: * & $\circ$ Always neat and organized \\
		& $\circ$ Neat most of the time \\
		& $\circ$ Cluttered most of the time \\
		& $\circ$ Always messy and disorganized \\
		& \\
		$\bullet$ What percentage of your floor is currently covered with stuff? & $\%$ $\square$  \\
		(e.g. clothes, books, papers, random other things)  &  \\
		& \\
		$\bullet$ I understand that roommates share a small space & $\circ$ Not comfortable sharing my stuff \\
		 and often choose to share certain items. In general, I am: * & $\circ$ Willing to share certain items only \\
		 & $\circ$ Willing to share most stuff if I'm asked first\\
		 & $\circ$ What's mine is yours and vice versa\\
		\hline
		$\bullet$ Which area is most important? & $\circ$ Sleep \\ Select TWO * & $\circ$ Music \\  & $\circ$ Sharing \\ & $\circ$ Cleanliness \\
		& $\circ$ Personality  \\
		\hline\hline
	\end{tabular}}
	\vspace{-\baselineskip}
    \end{minipage}
\end{table}

With these motivations, we extend \srtiasp to include such personal preferences of applicants. We call this extension \Personalizedsrtiasp.

\Personalizedsrtiasp considers an aggregate preference list $\prec''_{x}$ defined over two types of preference lists: $\prec_{x}$ as defined in the previous section, and $\prec'_{x}$ to capture the additional preferences as discussed above.

\paragraph{\bf Defining the criteria-based personalized preference lists $\prec'_{x}$.}
Let us first introduce some definitions and notations as follows.

Let $B$ be a finite list $\langle b_{1},b_{2},\dots,b_{k} \rangle$ of criteria. For each criterion $b_{i} \in B$, let $C_i$ be a finite list $\langle c_{i1},c_{i2},\dots,c_{im} \rangle$ of choices for $b_{i}$, that is ordered with respect to a ``closeness'' measure (i.e., for every choice $c_{ij}$, the choice $c_{ij'}$ ($j<j'$) is ``closer'' than the choice $c_{ij''}$ ($j'<j''$)). The closeness measure is useful for matching agents with closer choices, as roommates.
 For instance, consider the criteria list $B=\langle$``cleanliness", ``sleep habits"$\rangle$. For each criterion, the choice lists can be defined as follows: $C_1=\langle$``Clean", ``Messy"$\rangle$ is the list of choices for ``cleanliness", and $C_2=\langle$``Goes to bed early", ``Goes to bed before midnight", ``Goes to bed after midnight"$\rangle$ is the list of choices for ``sleep habits".

\begin{table}[t]
\caption{A personalized-\srti instance defined over an agent set $A=\{\text{Ayse, Buse, Cem, Duru}\}$, a criteria list $B=\langle$``smoking", ``cleanliness", ``room environment", ``sleep habits", ``study habits"$\rangle$, and the following choice lists for each criterion, $C_{1}=\langle$``Smoker",``Non-smoker"$\rangle$, $C_{2}=\langle$``Clean",``Messy"$\rangle$, $C_{3}=\langle$``Quiet",``Social",``Social and quiet"$\rangle$, $C_{4}=\langle$``Goes to bed early",``Goes to bed before midnight",``Goes to bed after midnight$\rangle$, $C_{5}=\langle$``Studies in the room",``Studies out of the room",``Studies in and out of the room"$\rangle$.}
\label{instance}
\vspace{-0.3\baselineskip}
\begin{minipage}{\textwidth}
\centering
\resizebox{\textwidth}{!}{\begin{tabular}{ccccc}
\hline\hline
Agent $x$  & Preference list $\prec_{x}$ & Profile $P_{x}$ & Weight list $W_{x}$ &  Extended preference list $\prec_{x}^{''}$ \\
\hline
Ayse & $\langle$Duru$\rangle$            & $\langle$2, 1, 1, 1, 1$\rangle$ & $\langle$5,4,3,2,1$\rangle$ & $\langle$Duru, Cem$\rangle$\\
(non-smoker) &           &   & &\\
\hline
Buse &       $\langle \rangle$       & $\langle$1, 2, 3, 3, 3$\rangle$ & $\langle$1,0,3,4,5$\rangle$ & $\langle$Duru, Cem$\rangle$  \\
(smoker) &             &   & & \\
\hline
Cem  & $\langle$Ayse, Buse$\rangle$ & $\langle$2, 1, 3, 2, 3$\rangle$ & $\langle$5,5,4,3,2$\rangle$ & $\langle$Ayse, Buse, Duru$\rangle$ \\
(non-smoker) & &  & & \\
\hline
Duru  & $\langle$Cem$\rangle$     &  $\langle$2, 1, 3, 3, 3$\rangle$ & $\langle$3,3,3,3,3$\rangle$ &  $\langle$Cem, Buse, Ayse$\rangle$ \\
(non-smoker) &     &  & &\\
\hline\hline
\end{tabular}}
\vspace{-\baselineskip}
\end{minipage}
\end{table}

Let $f$ be a function that maps an agent $x\in A$ and a criterion $b_{i} \in B$ to a positive integer $j$ ($1\leq j \leq |C_{i}|$), describing the choice $c_{ij}$ of the agent $x$. Consider the example above, and assume that Ayse is an agent in $A$. If Ayse's preference for the ``cleanliness" criterion is ``Clean", then $f($Ayse,``cleanliness"$)=1$. If Ayse's preference for ``sleep habits" criterion is ``Goes to bed after midnight", then $f($Ayse,``sleep habits"$)=3$.

For every agent $x\in A$, let us denote by $P_{x}=\langle f(x,b_{1}),f(x,b_{2}),\dots,f(x,b_{k})\rangle$ the choices of $x$ for each criterion in $B$ respectively. We refer to $P_{x}$ as the agent $x$'s (preference) profile. Consider the example shown in Table~\ref{instance}.
The preference profile $P_{Buse}$ for agent Buse is $\langle$1, 2, 3, 3, 3$\rangle$ where $B=\langle$``smoking", ``cleanliness", ``environment", ``sleep habits", ``study habits"$\rangle$. According to $P_{Buse}$, Buse prefers a roommate that is a ``Smoker'', `Messy'', ``Social and quiet'',``Goes to bed after midnight'',``Studies in and out of the room.''

Every criterion in $B$ may have a different importance for each agent. For instance, agent Ayse may give more importance to ``study habits'' while agent Buse gives more importance to ``cleanliness.'' To take into account the importance of these criteria, we introduce a weight function $w$ that maps an agent $x\in A$
and a criterion $b_{i} \in B$ to a non-negative integer such that $w(x,b_{i})$ denotes the importance of the criterion $b_{i}$ for $x\in A$. For every agent $x\in A$, let us denote by the weight list $W_{x}=\langle w(x,b_{1}),w(x,b_{2}),\dots,w(x,b_{k})\rangle$ the respective weights of criteria in $B$ for $x$. Note that $w(x,b_{i})>w(x,b_{j})$ implies that the criterion $b_{i}$ is more important than the criterion $b_{j}$ for agent~$x$. We say that $w(x,b_{i})=0$ to indicate that the criterion $b_{i}$ is not important for agent~$x$. For the example shown in Table~\ref{instance}, $W_{Buse}=\langle$1,0,3,4,5$\rangle$: the most important criterion for Buse is ``study habits", and the ``cleanliness" criterion is not important.

For every agent $x\in A$, with a profile $P_x$ and a weight list $W_x$,
let us denote the criteria of the same weight $u>0$
and the agent $x$'s choices for them, by a nonempty set $E_u$ of tuples as follows:
$$E_u = \{(f(x,\pi_{i}),\pi_i)\ |\ u=w(x,\pi_i)>0,\ \pi_i \in \{b_{1},b_{2},\dots,b_{k}\}\}.$$
Then, for every agent $x\in A$, we define a sorted profile $P_{x}^{'}$ for $x$,
with respect to $P_{x}$ and $W_x$,
as follows:
\begin{center}
$P_{x}^{'}=\langle E_{u_1}, E_{u_2}, \dots, E_{u_m} \rangle$
where $m \leq k$, and, for each $i$ ($1 {\leq} i {<} m$), $u_i > u_{i+1}$.
\end{center}
In Table~\ref{instance},
the sorted profile for Cem is
$P_{Cem}^{'} = \langle \{(2,$``smoking"$), (1,$``cleanliness"$)\},
\{(3,$``room environment"$)\},\ \{(2,$``sleep habits"$)\},\ \{(3,$``study habits"$)\}\rangle$
considering the importance of each criterion for him:
$w($Cem, ``smoking"$)=w($Cem, ``cleanliness"$)=5,
w($Cem, ``room$\\
$environment"$)=4,\ w($Cem,``sleep habits"$)=3,\ w($Cem,``study habits"$)=2.$

For every agent $y\in A \backslash A_{x}$ (i.e., $y$ is not acceptable to $x$),
if there exists some criterion $b_{i}\in B$ where $w(x,b_i)>0$ such that $f(y,b_{i})=f(x,b_{i})$, then we say that $y$ is {\em choice-acceptable} to $x$. We denote by $A_x'$ the set of all agents in $A \backslash A_{x}$ that are choice-acceptable for $x$. In Table~\ref{instance}, since Ayse has no common choice with Buse, Ayse is not choice-acceptable for Buse. On the other hand, Duru has a common choice with Buse: $f($Duru, ``study habits''$)= f($Buse, ``study habits''$) = 3$; and thus Duru is choice-acceptable for Buse. We assume that $x$ prefers every choice-acceptable $y$ as a roommate compared to being single.

For every agent $x$ with a sorted profile $P_{x}^{'}=\langle E_{u_1}, E_{u_2}, \dots, E_{u_m} \rangle$ ($m \leq k$), for every two agents $y$ and $z$ that are choice-acceptable to $x$, the agents $y$ and $z$ are {\em choice-equal} for $x$ relative to the first $j$ sets $E_{u_1}, E_{u_2}, \dots, E_{u_j}$ in $P_{x}^{'}$ (denoted $y =_{x} z |_j$) if the following holds:
\begin{itemize}
\item $j=0$, or
\item $j>0$, $y =_{x} z |_{j-1}$, and, for every $(f(x,\pi_{i}),\pi_i)\in E_{u_j}$, $f(x,\pi_{i}) = f(y,\pi_{i}) = f(z,\pi_{i})$.
\end{itemize}
We say that $x$ prefers $y$ to $z$ with respect to a sorted profile $P_{x}^{'}$ (denoted $y \prec'_{x} z$) if the following holds for some $j>0$:
\begin{itemize}
\item $y =_{x} z |_{j-1}$, and
\item $|\{\pi_i |\ (f(x,\pi_{i}),\pi_i){\in} E_{u_j},\ f(x,\pi_{i}) {=} f(y,\pi_{i})\}|\ >\
        |\{\pi_i |\ (f(x,\pi_{i}),\pi_i){\in} E_{u_j},\ f(x,\pi_{i}) {=} f(z,\pi_{i})\}|$.
\end{itemize}
For agent $x$, we say that $y \sim_{x}^{'} z$ if $y \not \prec_{x}^{'} z$ and $z \not \prec_{x}^{'} y$.

A {\em criteria-based personalized preference list} $\prec_{x}^{'}$ is a partial ordering of $x$'s preferences over~$A'_x$ with respect to a sorted profile $P_{x}^{'}$, where such incomparability is transitive.

For example, in Table~\ref{instance},
for Ayse,
$P_{Ayse}^{'}=\langle E_5, E_4, E_3, E_2, E_1\rangle$ where
$E_5=\{(2,$``smoking"$)\}$, $E_4=\{(1,$``cleanliness"$)\}$,
$E_3=\{(1,$``room environment"$)\}$, $E_2=\{(1,$``sleep habits"$)\}$, and $E_1=\{(1,$``study habits"$)\}$.
Cem is choice-acceptable for Ayse: $f(Cem,$``smoking"$){=}f($Ayse,``smoking"$){=}2$. Then, the criteria-based personalized preference list $\prec_{Ayse}^{'}$ is $\langle Cem\rangle$: Ayse prefers Cem as a roommate compared to being single.

For Buse, $P_{Buse}^{'}=\langle E_5, E_4, E_3, E_1\rangle$ where
$E_5=\{(3,$``study habits"$)\}$, $E_4=\{(3,$``sleep habits"$)\}$,
$E_3=\{(3,$``room environment"$)\}$, $E_1=\{(1,$``smoking"$)\}$.
Since Ayse has no common choice with Buse, Ayse is not choice-acceptable for Buse.
On the other hand, Cem and Duru are choice-acceptable for Buse.
Then, Duru~$\prec_{Buse}^{'}$~Cem since
\begin{itemize}
\item
for the criterion $\pi_1=$``study habits" in $E_5$, $f($Buse,$\pi_{1}){=}f($Duru,$\pi_{1}){=}f($Cem,$\pi_{1}){=}3$, and thus Duru $=_{Buse}$ Cem $|_1$; and
 \item
for the criterion $\pi_2{=}$``sleep habits" in $E_4$, $f($Duru,$\pi_{2}){=}f($Buse,$\pi_{2}){=}3 $ while
$f($Buse,$\pi_{2}){\neq}$\\
$f($Cem$,\pi_{2}){=}2$. Therefore, $|\{\pi_i |\ (f($Buse$,\pi_{i}),\pi_i){\in} E_4, f($Duru$,\pi_{i}) = f($Buse,$\pi_{i})\}| {=} 1$ is larger than
$|\{\pi_i |\ (f(x,\pi_{i}),\pi_i){\in} E_{4},\ f($Buse,$\pi_{i}) = f($Cem,$\pi_{i})\}| {=} 0$.
\end{itemize}
Then, the criteria-based personalized preference list $\prec_{Buse}^{'}$ is $\langle$Duru, Cem$\rangle$.

For Duru,
$P_{Duru}^{'}=\langle E_3\rangle$ where
$E_3 = \{(3,$``study habits"$), (3,$``sleep habits"$), (3,$``room environ-\\
ment"$), (2,$``smoking"$), (1,$``cleanliness"$)\}$. Ayse and Buse are choice-acceptable for Duru. Then, Buse~$\prec_{Duru}^{'}$~Ayse since
$|\{\pi_i |\ (f($Duru$,\pi_{i}),\pi_i){\in} E_3,\ f($Duru$,\pi_{i}) = f($Buse$,\pi_{i})\}| {=} 3$ is larger than
$|\{\pi_i |\ (f(x,\pi_{i}),\pi_i){\in} E_{u_1},\ f($Duru,$\pi_{i}) = f($Ayse,$\pi_{i})\}| {=} 2$.
Then the criteria-based personalized preference list $\prec_{Duru}^{'}$ is $\langle$Buse, Ayse$\rangle$.

\paragraph{\bf Defining the extended preference lists $\prec''_{x}$.}
We define $\prec_{x}^{''}$ as an extended preference list by concatenating $\prec_{x}$ and  $\prec_{x}^{'}$ depending on the importance given to these two types of lists.
For the instance in Table~\ref{instance}, suppose that the preference lists $\prec_{x}$ are more important. Then the preference list $\prec_{x}^{'}$ is appended to end of the preference list $\prec_{x}$. Then
the extended preference list of Buse is $\prec_{Buse}^{''}=\langle$Duru, Cem$\rangle$. The extended preference lists for other agents are as shown in Table~\ref{instance}.

\paragraph{\bf Personalized-\srti} is then characterized by $(A,\prec^{''})$ where $A$ is a finite set of agent, and $\prec^{''}$ is collection of the extended preference list of each agent $x \in A$. To solve Personalized-\srti, we utilize \srtiasp as described in Section \ref{sec:srti}.


\section{Most-\srti: \srti with Most Preferred Criteria} \label{secondApproach}

Instead of considering individual importance of the criteria for each agent, we can consider the most preferred  criteria (e.g., identified by large surveys) and try to find stable roommate matchings accordingly. For such applications, we introduce a new  definition for stable matchings.

\paragraph{\bf Most-\srti.} Let $W$ be a criteria list $\langle b_{1},b_{2},\dots,b_{k} \rangle$ sorted with respect to their overall importance for all agents. For each criterion $b_{i} \in W$, let $C_i$ be a finite list of choices ordered with respect to a closeness measure, as discussed in the previous section. Let $f$ be a function that maps an agent $x\in A$ and a criterion $b_{i} \in W$ to a positive integer $j$ ($a \leq j \leq |C_{i}|$).

We start with the set $\mathcal{M}$ of all stable matchings of a given \srti instance $(A, \prec)$, and define a series of subsets $\mathcal{M}_{max}(i)$ of these matchings to maximize the overall satisfaction of the roommates with respect to the closeness of their choices for the criterion $b_{1},b_{2},\dots,b_{k}$:
$$
\ba l
\hspace{-1ex}\mathcal{M}_{max}(0) = \mathcal{M} \\
\hspace{-1ex}\mathcal{M}_{max}(i) = \{M {\in} \mathcal{M}_{max}(i{-}1) |\ 1 {\leq} i {\leq} |W|,\ \forall\ M' {\in} \mathcal{M}_{max}(i{-}1)\ \textrm{s.t.}\ M' {\neq} M,\\
\qquad \sum\limits_{x\in A} |f(x,b_{i})-f(M'(x),b_{i})| \geq \sum\limits_{x\in A} |f(x,b_{i})-f(M(x),b_{i})| \}.
\ea
$$
Then, a stable matching $M \in \mathcal{M}_{max}(|W|)$ is called a {\em  most preferred criteria based stable matching} with respect to the criteria list $W$. We call the problem of finding such a stable matching, Most-\srti.

For example, consider the instance in Table~\ref{instance}. Instead of considering the individual importance of the criteria for each applicant, let us take $W=\langle$``smoking", ``cleanliness", ``room environment", ``sleep habits", ``study habits"$\rangle$. Hence, we try to find a matching that maximizes first the number of roommates which are close to each others in terms of their smoking criteria, and then, subject to this condition, maximizes the number of roommates which are close to each other in terms of their cleanliness criteria, and then, subject to this condition, maximizes the number of roommates which are close to each others in terms of their room environment criteria, and then, subject to this condition, maximizes the number of roommates which are close to each others in terms of their sleep habits criteria, and then, subject to this condition, to maximizes the number of roommates which are close to each others in terms of their study habits criteria. A stable matching at the end is called a most-preferred stable matching.

\paragraph{\bf Solving Most-\srti using ASP}

We can solve Most-\srti in ASP utilizing weighted weak constraints of different priorities. The idea is to introduce weighted weak constraints to express preferences for each criterion, where the higher priorities are given for the most preferred criteria.

For each agent $x$, for each criterion $b_{i}\in W$, we describe the choice $r \in C_i$ of $x$ for $b_i$ (i.e., $f(x,b_i)=r$) by atoms. For instance, we introduce atoms of the form $\ii{bedTime}(x,r)$ to describe that $f(x,\text{``sleep habits"})=r$. Then the preferences of agents can be represented as follows:
\begin{itemize}
\item $\textit{bedTime}(x,1)$: the agent $x$ prefers a roommate who goes to bed before $11$ pm,
\item $\textit{bedTime}(x,2)$: the agent $x$ prefers a roommate who goes to bed before midnight,
\item $\textit{bedTime}(x,3)$: the agent $x$ prefers a roommate who goes to bed after midnight.
\end{itemize}
Using these atoms, the following weak constraint tries to maximize the number of roommates who are close to each other in terms of their sleep habits:
\beq
\mathrel{\mathop{\lar}^{\sim}} \{\ii{room}(x,y), \ii{bedTime}(x,r1), \ii{bedTime}(y,r2)\}.[|r1-r2|@p,x,y]
\eeq{eq:sleepHabits}
\noindent Here, the priority $p$ is assigned a high value if ``sleep habits" is one of the most preferred criteria.

For the ``cleanliness" criterion, the preferences of agents can be represented by the following atoms of the forms:
\begin{itemize}
\item $\textit{cleanliness}(x,1)$: the agent $x$ tends to keep his/her room clean,
\item $\textit{cleanliness}(x,2)$: the agent $x$ tends to keep his/her room messy.
\end{itemize}
Using these atoms, the following weak constraints try to maximize the number of roommates who are close to each other in terms of their cleanliness degrees:
\beq
\begin{aligned}
\mathrel{\mathop{\lar}^{\sim}}& \{\ii{room}(x,y), \ii{cleanliness}(x,r1), \ii{cleanliness}(y,r2)\}.[|r1-r2|@p,x,y] \\
\end{aligned}
\eeq{eq:cleanliness}

Consider, for instance, ``smoking" habits.
This is an important criterion to match roommates even if they live on a smoke-free campus. According to the following questions:\footnotemark[\getrefnumber{cleveland}]
\begin{itemize}
\item Are you smoker?  $\circ$ Yes  $\circ$ No
\item Are you comfortable with a roommate that is a smoker?  $\circ$ Yes  $\circ$ No
\end{itemize}
we can describe the smoking habits of the agents with atoms of the forms $\ii{smoker}(x)$,  $\ii{nonsmoker}(x)$, and their preferences with the following atoms of the forms:

\begin{itemize}
\item $\textit{comfortableSmoker}(x,1)$: the agent $x$ is comfortable with a smoker roommate,
\item $\textit{comfortableSmoker}(x,2)$: the agent $x$ is not comfortable with a smoker roommate.
\end{itemize}
We can define non-smoker agents who are comfortable with a smoker roommate:
$$ 
\ii{smokeComfor}(x,y) \mathrel{\mathop{\lar}} \ii{nonsmoker}(x), \ii{comfortableSmoker}(x,1), \ii{smoker}(y).
$$ 
We can define agents who is not comfortable with a smoker roommate:
$$ 
\neg \ii{smokeComfor}(x,y) \mathrel{\mathop{\lar}} \ii{comfortableSmoker}(x,2), \ii{smoker}(y).
$$ 
Then the following weak constraints can be added to our ASP formulation to maximize the number of roommates who are comfortable with each others in terms of their smoking habits with the given priority $p$:
\beq
\begin{aligned}
\mathrel{\mathop{\lar}^{\sim}}& \{\ii{room}(x,y), \text{not}~\ii{smokeComfor}(x,y), \ii{nonsmoker}(x), \ii{smoker}(y)\}.[1@p,x,y] \\
\mathrel{\mathop{\lar}^{\sim}}& \{\ii{room}(x,y), \neg \ii{smokeComfor}(x,y), \ii{smoker}(x), \ii{smoker}(y)\}.[1@p,x,y]
\end{aligned}
\eeq{eq:smokingHabits}

According to the ``room Environment'' criterion,\footnotemark[\getrefnumber{clark}] the preferences of agents can be represented by atoms of the form:
\begin{itemize}
\item $\textit{roomEnvironment}(x,1)$: the agent $x$ wants his/her room to be quiet and study oriented,
\item $\textit{roomEnvironment}(x,2)$: the agent $x$ wants his/her room to social gathering place for friends to hang out,
\item $\textit{roomEnvironment}(x,3)$: the agent $x$ wants his/her room to be a combination of social and quiet.
\end{itemize}
Using these atoms, the following weak constraint tries to maximize the number of roommates who are close to each other in terms of their room description:
\beq
\mathrel{\mathop{\lar}^{\sim}} \{\ii{room}(x,y), \ii{roomEnvironment}(x,r1), \ii{roomEnvironment}(y,r2)\}.[|r1-r2|@p,x,y]
\eeq{eq:roomEnvironment}

Another important criterion is ``study Habits.'' For this criterion,\footnotemark[\getrefnumber{clark}]
the preferences of applicants can be represented by the following atoms of the form $\ii{studyHabit}(x,r)$:
\begin{itemize}
\item $\ii{studyHabit}(x,1)$: the agent $x$ expects to study in his/her room,
\item $\ii{studyHabit}(x,2)$: the agent $x$ expects to study outside of his/her room,
\item $\ii{studyHabit}(x,3)$: the agent $x$ expects to study both inside and outside of his/her room.
\end{itemize}
Using these atoms, the following weak constraint tries to maximize the number of roommates which are close to each other in terms of their study environment:
\beq
\mathrel{\mathop{\lar}^{\sim}} \{\ii{room}(x,y), \ii{studyHabit}(x,r1), \ii{studyHabit}(y,r2)\}.[|r1-r2|@p',x,y]
\eeq{eq:studyHabits}
\noindent Here, the priority $p'$ is assigned a lower value if {\em Study Habits} is not one of the most preferred criteria.

Note that we can combine different domain-independent measures of \srti with domain-specific measures, by assigning different priorities to them.


\section{Diversity Preferences}

In addition to the student's preferences, the schools may prefer matchings to increase diversity. For example, they may want to match student from various departments, different classes, countries. Also, some students may be forbidden to match with each other (like in the hedonic diversity games~\cite{boehmer2020}) where the school partition the students into two groups for diversity preferences.

Consider, for instance, maximizing the number of roommates from different departments at a university. A student's department can be defined by atoms of the form $\ii{department}(x,d)$ (``the student $x$'s department is $d$''). Then, the following weak constraints can be added to our ASP formulation, to maximize the number of roommates from different departments:
$$ 
\mathrel{\mathop{\lar}^{\sim}} \{\ii{room}(x,y), \ii{department}(x,d1), \ii{department}(y,d2),~d1 \neq d2 \}.[1@p,x,y]
$$ %

The school may not want to allow some students to be roommates. Then, such students can be defined by atoms of the form $\ii{forbidden}(x,y)$ (``students $x$ and $y$ are forbidden to be roommates''), and the following hard constraints can be added to our ASP formulation:
$$ 
\lar \ii{forbidden}(x,y),~\ii{room}(x,y) \qquad (x\neq y).
$$

Therefore, the diversity-related constraints and preferences can be easily added to \srtiasp.


\section{Experimental Evaluations}

We have experimentally evaluated Personalized-\srti to understand its scalability over \srti instances with additional knowledge, and compared Personalized-\srti with Most-\srti.

\paragraph{\bf Scalability of Personalized-\srti.}
For benchmarks, as a basis, we have used the \srti instances randomly generated for our earlier experiments~\cite{erdem2020}. It is based on the following idea~\cite{Mertens2005}: 1) generate a random graph ensemble $G(n,p)$ according to the Erdos-Renyi model~\cite{Erdos60}, where $n$ is the required number of agents and $p$ is the edge probability (i.e., each pair of vertices is connected independently with probability $p$); 2) since the edges characterize the acceptability relations, generate a random permutation of each agent’s acceptable partners to provide the preference lists. We define the {\em completeness degree} for an instance as the percentage $p*100$.

We have considered instances of different sizes, where the number of agents are 40, 60, 80,100, 150 and 200, and the completeness degrees are 25\%, 50\%.  For each number of agents and for each completeness degree, there are 20 instances.  Then, for each instance, for each agent in that instance, we have randomly generated the agent's choices for each criterion, and the importance of each given criteria according to the agent. For each instance, we have considered 2--5 criteria.

\begin{figure}
 \centering
    \begin{tabular}{cc}
	\includegraphics[width=0.5\linewidth]{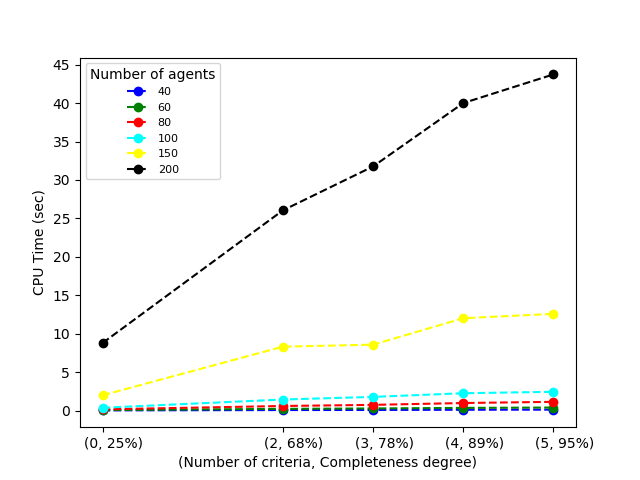}
	&
	\includegraphics[width=0.5\linewidth]{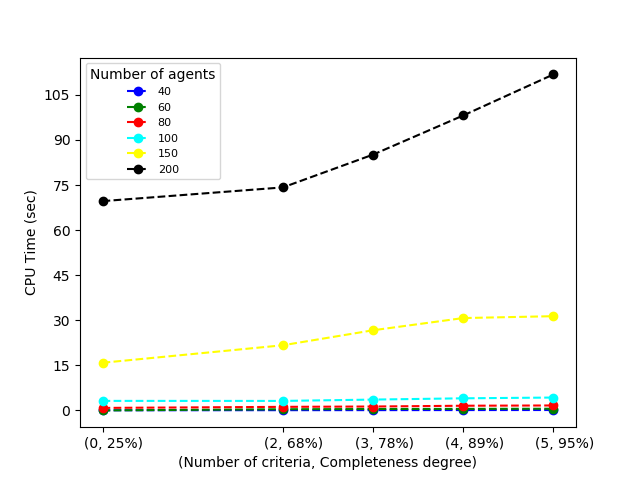}
    \end{tabular}
    \vspace{-0.3\baselineskip}
	\caption{Scalability of \Personalizedsrtiasp as the number of criteria and the completeness degrees increase, for instances where the initial completeness degree is 25\% (the left figure) and the initial completeness degree is 50\% (the right figure).}
	\label{p-srti}
\vspace{-0.5\baselineskip}
\end{figure}

In our experiments, we have used \clingo (Version 5.2.2) on a machine with Intel Xeon(R) W-2155 3.30GHz CPU and 32GB RAM. The results are shown in Figure~\ref{p-srti}.

We make the following observations from this figure, similar to our observations~\cite{erdem2020} over \srti experiments:
As the number of agents and the completeness degree increase, the computation times increase. In addition, as the number of criteria increase, the computation times increase.

Note that the initial completeness degree changes as additional knowledge is included about preferences of agents over different criteria. For Personalized-\srti instance, the completeness degree is around
$$
\dfrac{n\times \dfrac{d}{100}+ n\times (1-\dfrac{d}{100})\times\dfrac{m}{m+1}}{n} \times 100
$$
where $n$ is the number of agents, $d$ is the initial completeness degree, and $m$ is the number of criteria. Therefore, the completeness degree of a Personalized-\srti depends on the initial completeness degree and the number of criteria but not on the number of agents. Consider an instance where $d=25$ and $m=3$. We expect that the completeness degree be around $81\%$ depending on the preferences of the agents. In fact, the completeness degree in our experiments is~$83\%$ (Figure~\ref{p-srti}).

\paragraph{\bf Personalized-\srti vs. Most-\srti.}
For benchmarks, as a basis, we have used the \srti instances randomly generated for our earlier experiments~\cite{erdem2020} as described above.

For each instance, for each agent, we have randomly generated the agent's choices for the most popular three criteria, which are cleanliness, sleep habit and study habit.  The importance of each given criteria is fixed as $1, 2, 3$ respectively.

The results of our experiments for 40--200 agents are shown in Table~\ref{tab:kb1-kb2}. We can observe that Most-\srti performs better than Personalized-\srti. For both approaches, the computation times for finding a stable matching (if one exists) and finding out that there exists no stable matching are comparable to each other. We can make further observation: As the completeness degree increase, the computation times of Most-\srti more increase than Personalized-\srti.

\begin{table}[t]
\caption{Personalized-\srti vs Most-\srti.}
\label{tab:kb1-kb2}
\vspace{-0.3\baselineskip}
\begin{minipage}{\textwidth}
\centering
\resizebox{\textwidth}{!}{\begin{tabular}{cccccc}
\hline \hline
            &    	& &  \multicolumn{2}{c}{Personalized-\srti}  & \multicolumn{1}{c}{Most-\srti}   \\
initial completeness&  		 & \#instances with& completeness degree & avg. time    & avg. time        \\
degree       & $|A|$ & a solution & with additional knowledge &(sec)   & (sec)    \\
   \hline
25\%
 & 40 & 11 & 83\% & 0.109 & 0.015  \\
 & 60 & 10 & & 0.379  & 0.056   \\
 & 80 & 13 & & 1.072  & 0.142 \\
 & 100 & 14 & & 2.469 & 0.353 \\
 & 150 & 8 & & 14.556  & 2.331  \\
 & 200 & 10 & & 51.640  & 11.116 \\
\hline
50\%
 & 40 & 11 & 89\% & 0.149  & 0.059 \\
 & 60 & 16 &  & 0.636  & 0.257 \\
 & 80 & 13 &  & 1.903  & 0.822 \\
 & 100 & 12 &  & 4.610  & 2.497  \\
 & 150 & 14 &  & 42.273  & 25.563  \\
 & 200 & 9 & & 140.378 & 104.921  \\
 \hline \hline
\end{tabular}}
 \vspace{-\baselineskip}
\end{minipage}
\end{table}


\section{A Real-World Application}

In collaboration with more than 200 students at Sabanci University, we have investigated the applicability of our methods for Personalized-\srti.

First, we have conducted a survey to select the most important 5 criteria that should be included in a dormitory application. Next, we have conducted a survey to get the preferences of each student for each criterion. Next, we have conducted a survey to evaluate the usefulness of Personalized-\srti from the perspective of students. The surveys are given several months apart from each other, considering the availabilities of students.

\subsection{First Survey: Which criteria should be considered in a dormitory questionnaire?}

Students prefer short application forms. With this motivation, first we have conducted a survey to find out which multiple choice questions presented in the second part of Table~\ref{firstsurvey} should be included in a dormitory questionnaire.

We have conducted this survey online (due to pandemic), at Sabanci University: 156 students have participated, 120 of them live in the dormitories, and 36 of them do not.

\begin{figure}[t]
  \centering
  \includegraphics[width=0.98\linewidth]{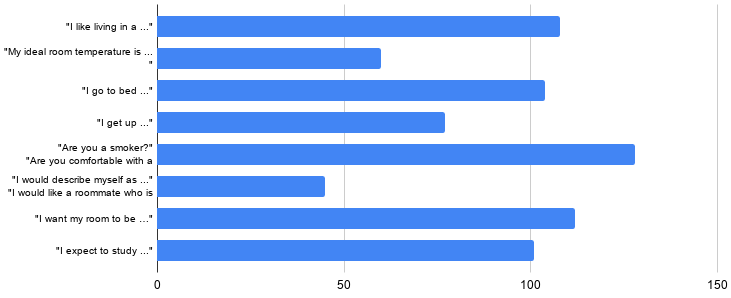}
	\caption{Results of the First Survey}
	\label{barchart}
\end{figure}

Figure~\ref{barchart} shows the most popular five questions, chosen by more than 100 students.

\begin{table}[t]
\caption{Roommate Questionnaire (First Survey)}
\label{firstsurvey}
\vspace{-0.3\baselineskip}
\begin{minipage}{\textwidth}
\centering
\resizebox{0.8\textwidth}{!}{%
\begin{tabular}{l}
\hline \hline
Part 1: About Yourself \\
\quad 1. Do you live in dormitory? \\
\qquad $\circ$ Yes \\
\qquad $\circ$ No \\
\quad 2. Which one do you prefer? \\
\qquad $\circ$ Random roommate \\
\qquad $\circ$ Roommate with similar expectations \\ \\
Part 2: About Dormitory Questionnaires \\
\quad 3. Which multiple choice questions should be included in the roommate questionnaire? \\
\quad $\circ$ "I like living in a ..." (a) Clean Space   (b) Messy Space   (c) Indifferent \\
\quad  $\circ$ "My ideal room temperature is ... "    (a) Cold (below 18 °C)    (b) Fairly cold (18 °C-21 °C)\\
\quad \qquad \qquad \qquad  \qquad \qquad \qquad \qquad \quad (c) Fairly warm (21 °C-24 °C)    (d) Warm (above 24 °C) \\
\quad $\circ$ "I go to bed ..."     (a) Before 11pm    (b) Before Midnight    (c) After Midnight \\
\quad $\circ$ "I get up ..." (a) Before 8am (b) 8am - 10am (c) 10am - 12pm (d) 12pm or later \\
\quad $\circ$ "Are you a smoker?" \&  "Are you comfortable with a roommate that is a smoker?"  (a) Yes (b) No \\
\quad $\circ$ "I would describe myself as ..."  \&  "I would like a roommate who is ..." (a) Shy (b) Fairly Shy \\
\quad \qquad \qquad \qquad  \qquad \qquad \qquad \qquad  \qquad \qquad \qquad  \qquad \qquad \qquad  \qquad \quad ~(c) Fairly Outgoing (d) Outgoing \\
\quad $\circ$ "I want my room to be ..." (a) Quiet and study oriented (b) A social gathering place for friends to hang out \\
\quad \qquad \qquad \qquad \qquad \qquad \quad ~~(c) A combination of social and quiet \\
\quad $\circ$ "I expect to study ..."  (a) In my room    (b) Outside of my room    (c) Both inside and outside of my room \\ \hline \hline
\end{tabular}%
}
\vspace{-\baselineskip}
\end{minipage}
\end{table}

\begin{table}[]
\caption{Roommate Questionnaire (Second Survey)}
\label{tab:my-form}
\vspace{-0.3\baselineskip}
\begin{minipage}{\textwidth}
\centering
\resizebox{0.8\textwidth}{!}{%
\begin{tabular}{p{0.6\textwidth}p{0.35\textwidth}}
\hline \hline
First Name : &  \\
Last Name : &  \\
Email : &  \\
Gender : & $\circ$ Female $\circ$ Male \\
I am requesting a specific roommate: &  \\ \hline
\multicolumn{2}{l}{{\ul Sleep Habits:}} \\
\multirow{3}{*}{$\bullet$ I go to bed ...} & $\circ$ Before 11pm \\
 & $\circ$  Before Midnight \\
 & $\circ$  After Midnight \\ \hline
\multicolumn{2}{l}{{\ul Cleanliness:}} \\
\multirow{3}{*}{$\bullet$ I like living in a ...} & $\circ$ Clean Place \\
 & $\circ$  Messy Place \\
 & $\circ$  Indifferent \\ \hline
\multicolumn{2}{l}{{\ul Smoking Habits:}} \\
\multirow{2}{*}{$\bullet$ Are you a smoker?} & $\circ$  Yes \\
 & $\circ$  No \\ &  \\
\multirow{2}{*}{$\bullet$ Are you comfortable with a roommate that is a smoker?} & $\circ$  Yes \\
 & $\circ$  No \\ \hline
\multicolumn{2}{l}{{\ul Room Environment:}} \\
\multirow{3}{*}{$\bullet$ I want my room to be ...} & $\circ$  Quiet and study oriented \\
 & $\circ$  A social gathering place for friends to hang out \\
 & $\circ$  A combination of social and quiet \\ \hline
\multicolumn{2}{l}{{\ul Study Habits:}} \\
\multirow{3}{*}{$\bullet$ I expect to study ...} & $\circ$ In my room \\
 & $\circ$  Outside of my room \\
 & $\circ$  Both inside and outside of my room \\ \hline

Indicate the importance on the scale 1-5 of each of the following & \\
with (1) being very important to you and  &\\
		\qquad (5) being of little importance to you:& \\
		\textbf{Sleep Habits: } \underline{\hspace{1cm}}&  \\
		\textbf{Cleanliness: } \underline{\hspace{1cm}} &\\
		\textbf{Smoking Habits: } \underline{\hspace{1cm}}&  \\
		\textbf{Room Environment: } \underline{\hspace{1cm}}&  \\
		\textbf{Study Habits: } \underline{\hspace{1cm}} &\\
		\hline \hline		
\end{tabular}}
\vspace{-\baselineskip}
\end{minipage}
\end{table}

\subsection{Second Survey: What are your preferences?}

As a result of the first survey in Table~\ref{firstsurvey}, a roommate questionnaire (Table~\ref{tab:my-form}) is prepared with respect to the most preferred five criteria. The purpose of this survey is to generate real data for roommate matching: for each student, we gather the importance of criteria as well as their preferences for each criterion.

We have conducted this survey online (due to pandemic), at Sabanci University:  81 students have filled this survey.

According to the survey results, the following order of the given criteria describes the overall importance: smoking habits, cleanliness, room environment, sleep habits, and study habits.

This suggests solving a Most-\srti problem instance, where the goal is to find a most preferred criteria based stable matching that tries to maximize first the number of roommates which are comfortable with each others in terms of their smoking habits, and then, subject to this condition,
the number of roommates which are close to each others in terms of their cleanliness degree, and then, subject to this condition, the number of roommates which are close to each others in terms of their room description, and then, subject to this condition, the number of roommates which are prefer the same bedtime as close as possible, and then, subject to this condition, the number of roommates which are close to each others in terms of their study environment.

As described in Section~\ref{secondApproach}, we add weighted weak constraints to our ASP formulation of \srti (as described in Section~\ref{sec:srti}) to express preferences for each one of these five criteria, where the higher priorities are given for the most preferred criteria. Since the most important criteria is smoking habits, we add a weak constraint~(\ref{eq:smokingHabits}) where the priority $p$ is 5. Then, the next important criteria is cleanliness, we add a weak constraint~(\ref{eq:cleanliness}) where the priority $p$ is 4. Then, we add a weak constraint~(\ref{eq:roomEnvironment}) where the priority $p$ is 3. Then, we add a weak constraint~(\ref{eq:sleepHabits}) where the priority $p$ is 2. Finally, we add a weak constraint~(\ref{eq:studyHabits}) where the priority $p$ is 1.

\begin{figure}[t]
\centering
  \includegraphics[width=0.6\linewidth]{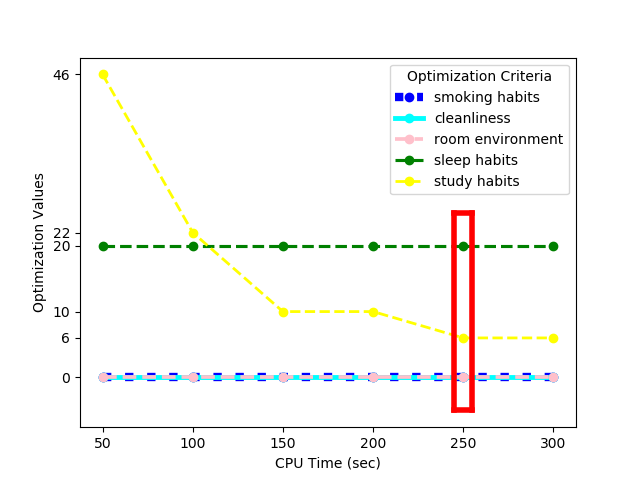}
  \vspace{-0.3\baselineskip}
	\caption{Computation of an optimal solution with anytime search.}
	\label{rectangle}
\vspace{-\baselineskip}
\end{figure}

Using our ASP program augmented with all these weak constraints, we have experimented over the real data  collected in this survey (i.e., preferences of agents for each criterion).  A most preferred criteria based stable matching is computed in 4403.51 seconds, where roommates are comfortable to each others in terms of the first three optimization criteria (smoking habits, cleanliness, room environment) but the importance of the sleep and study habits are 20 and 6 respectively. With anytime search, Figure~\ref{rectangle} shows that the most preferred criteria based stable matching is actually computed in 250 seconds; so the rest of the time is spent for optimality check.

\subsection{Third Survey: How good are the results of \Personalizedsrtiasp compared to unstable matchings?}\label{thirdSurvey}

\begin{figure}[t]
\centering
\begin{tabular}{cc}
  \includegraphics[width=0.43\linewidth]{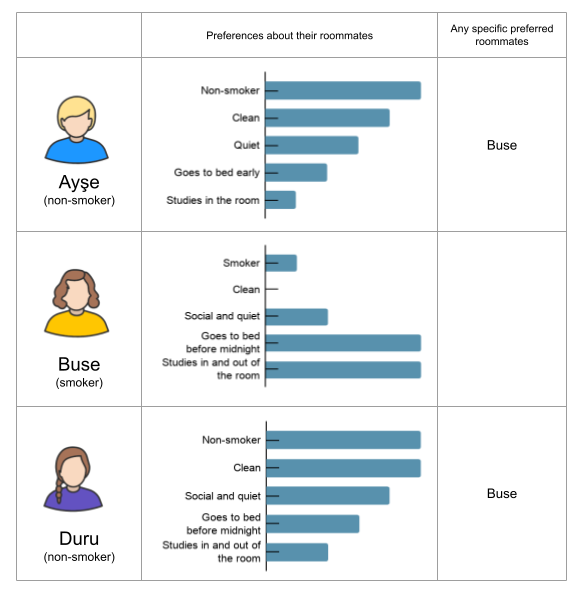}
  & \includegraphics[width=0.45\linewidth]{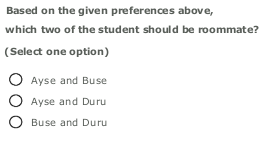}
\end{tabular}
\vspace{-0.3\baselineskip}
\caption{An example question of the third survey: Given the preferences shown on the left hand side, choose the most reasonable roommates on the right hand side.}
\label{eval-survey}
\vspace{-\baselineskip}
\end{figure}

In this survey, we have presented to the participants 3 Personalized-\srti instances with 3 agents (like in Figure~\ref{eval-survey}). Each instance is presented with 3 matchings, including a personalized stable matching computed by \Personalizedsrtiasp\ and 2 unstable matchings. We have requested the participants to choose the matching that makes sense the most. If they choose a matching different from the one computed by our method, we have asked for an explanation.

We have conducted this survey online (due to pandemic), at Sabanci University:  59  students have participated in this survey. The survey is conducted in three groups (Red, Blue, Green) with different orderings of instances.

\begin{table}[b]
\vspace{-\baselineskip}
\centering
\caption{Results of the Third Survey: Percentages of the participants who chose the personalized stable matching computed by \Personalizedsrtiasp. }
\label{percentage-survey}
\vspace{-0.3\baselineskip}
\begin{minipage}{0.95\textwidth}
\resizebox{0.95\textwidth}{!}{\begin{tabular}{cccc}
\hline\hline
Participants (\#) & Question 1 & Question 2 & Question 3\\
\hline
Red Group (13)      & 77\%  & 15\%  & 15\%        \\
Blue Group (24)     & 17\%  & 96\%  & 79\%  \\
Green Group (22) & 100\%   & 73\%  & 27\% \\
\hline\hline
\end{tabular}}
\end{minipage}
\vspace{-\baselineskip}
\end{table}

The percentages of choosing the stable matching computed by our method is shown in Table~\ref{percentage-survey}. According to the results, for 3 questions, many participants have chosen the personalized stable matchings computed by \Personalizedsrtiasp.
This shows that extending the preferences of agents with additional information about their habits and room environments is useful for the stable roommates problems.

We have also made interesting observations from the feedback and explanations provided by the participants, when they choose a matching different from the one computed by our method (over the remaining 2 instances). For instance, for the question shown in Figure~\ref{eval-survey} (Red Group, Question 2), although both Ay\c{s}e and Duru stated that they want Buse as their roommate, 85\% of the survey respondents chose Ay\c{s}e and Duru as the best roommate pair based on the given preferences. Eight of these participants stated that the reason why they chose Ay\c{s}e and Duru is that ``They give more importance to both smoking habits and cleanliness habits", one of them stated that ``They both prefer non-smoker roommates", and one of them specified the reason as ``only cleaning matters".  This feedback shows that the participants focus more on the additional information about habits and room environments, rather than specific preferences of roommates. In that sense, extending the preferences of agents with such additional information is useful. Furthermore, these results show that the participants have also considered their own preferences and priorities while choosing the best roommates.

\subsection{Fourth Survey: How good are the results of \Personalizedsrtiasp compared to the results of \srtiasp?}

In this survey, we have presented to the participants 3 Personalized-\srti instances (like in Section~\ref{thirdSurvey}). These instances consider 4 agents. Each instance  comes with 3 matchings to choose from: a personalized stable matching computed by \Personalizedsrtiasp, a stable matching computed by \srtiasp, and an unstable matching. We have requested the participants to choose the matching that makes sense the most.

We have conducted this survey online (due to pandemic), at Sabanci University:  42  students have participated in this survey. The survey is conducted in two groups (Blue, Green) with different orderings of instances.

According to the results (Table~\ref{percentage-survey4}),
while the overall percentage of choosing the stable matchings computed by \srtiasp is $29\%$, the overall percentage of choosing the personalized stable matchings computed by \Personalizedsrtiasp is $49\%$. These results illustrate that extending the preferences of agents with additional information about their habits and room environments is useful.

\begin{table}[b]
\vspace{-\baselineskip}
\centering
\caption{Results of the Fourth Survey: Percentages of the participants who chose the personalized stable matching computed by \Personalizedsrtiasp. }
\label{percentage-survey4}
\vspace{-0.3\baselineskip}
\begin{minipage}{0.95\textwidth}
\resizebox{0.95\textwidth}{!}{\begin{tabular}{cccc}
\hline\hline
                  & personalized stable matching    &   stable matching   & unstable matching\\
Participants (\#) & (\Personalizedsrtiasp)          &   (\srtiasp)        & \\
\hline
Green Group (13) & 46\%   & 39\%  & 15\% \\
Blue Group (29) & 50\%  & 25\%  & 25\%  \\
\hline
Overall (42) & 49\%  & 29\% & 22\% \\
\hline\hline
\end{tabular}}
\end{minipage}
\vspace{-\baselineskip}
\end{table}

\section{Conclusion}

We have extended \srtiasp to consider domain-specific knowledge about each individual's preferences about a set of criteria (e.g., about the habits of their roommates and the room environments), and about the diversity preferences of dormitories and schools (e.g., for assigning roommates from different departments).
We have in particular introduced two methods taking into account these additional preferences. Personalized-\srti considers personal preferences for each criterion, and the importance of the criteria for each agent, while Most-\srti considers personal preferences for the most preferred criteria (e.g., obtained by a survey as in our application).

We have also evaluated \Personalizedsrtiasp over different sizes of randomly generated Personalized-\srti instances, and compared it with Most-\srtiasp. We have observed that, although \Personalizedsrtiasp pays more attention to individuals' preferences, Most-\srtiasp performs better in computation time.

We have illustrated a real-world application of \Personalizedsrtiasp by interacting with at least 200 students at Sabanci University.
First, we have conducted a survey to select the most important five criteria that should be included in a dormitory application. Next, we have conducted a survey to get the preferences of each student for each criterion; in this way we have also collected real data for our experiments. Next, we have conducted two surveys to evaluate the usefulness of Personalized-\srti from the perspective of students. We have observed that many participants have chosen the solutions computed by \Personalizedsrtiasp,
and have given more importance to the additional information about habits and room environments. In that sense, extending \srti to include additional domain-specific knowledge is useful.

\paragraph{\bf Acknowledgments.}We would like to thank Mustafa Oguz Afacan and Selin Eyupoglu for useful discussions, and anonymous reviewers for their valuable comments. We would also like to thank the participants of the surveys.

\bibliographystyle{acmtrans}

\end{document}